\documentclass[runningheads]{llncs}

\usepackage{graphicx}
\usepackage[hidelinks]{hyperref}
\usepackage{lscape} 
\usepackage{multicol}
\usepackage{afterpage}
\usepackage{float}
\begin{document}

\title{Supervised Ontology and Instance Matching with MELT} %

\author{Sven Hertling\inst{1}\thanks{The authors contributed equally to this paper.} \orcidID{0000-0003-0333-5888} \and 
Jan Portisch\inst{1,2}$^\star$ \orcidID{0000-0001-5420-0663} \and
Heiko Paulheim\inst{1}\orcidID{0000-0003-4386-8195}}

\authorrunning{Sven Hertling, Jan Portisch, and Heiko Paulheim}

\institute{Data and Web Science Group, University of Mannheim, Germany\\
	\email{\{jan, sven, heiko\}@informatik.uni-mannheim.de} \and
	SAP SE Product Engineering Financial Services, Walldorf, Germany\\
\email{\{jan.portisch\}@sap.com}	
}

\maketitle 

\begin{abstract}
In this paper, we present \emph{MELT-ML}, a machine learning extension to the \emph{Matching and EvaLuation Toolkit} (MELT) which facilitates the application of supervised learning for ontology and instance matching. Our contributions are twofold: We present an open source machine learning extension to the matching toolkit as well as two supervised learning use cases demonstrating the capabilities of the new extension.

\end{abstract}

{\renewcommand\thefootnote{}\footnote{
Copyright © 2020 for this paper by its authors. Use permitted under Creative Commons License Attribution 4.0 International (CC BY 4.0).
}\addtocounter{footnote}{-1}}

\section{Introduction}
Many similarity metrics and matching approaches have been proposed and developed up to date. They are typically implemented as engineered systems which apply a process-oriented matching pipeline. Manually combining metrics, also called \emph{features} in the machine learning jargon, is typically very cumbersome. Supervised learning allows researchers and developers to focus on adding and defining features and to leave the weighting of those and the decision making to a machine. This approach may also be suitable for developing generic matching systems that self-adapt depending on specific datasets or domains. Here, it makes sense to test and evaluate multiple classifiers at once in a fair, i.e. reproducible, way. 
Furthermore, recent advances in machine learning -- such as in the area of knowledge graph embeddings -- may also be applicable for the ontology and instance matching community. The existing evaluation and development platforms, such as the \emph{Alignment API}~\cite{alignment_api_reference}, \emph{SEALS}~\cite{garcia2010towards,seals_2} or the \emph{HOBBIT}~\cite{ngomo2016hobbit} framework, make the application of such advances not as simple as it could be. 

In this paper, we present \emph{MELT-ML}, an extension to the \emph{Matching and EvaLuation Toolkit} (MELT). Our contribution is twofold: Firstly, we present a machine learning extension to the MELT framework (available in MELT 2.6) which simplifies the application of advanced machine learning algorithms in matching systems and which helps researchers to evaluate systems that exploit such techniques. Secondly, we present and evaluate two novel approaches in an exemplary manner implemented and evaluated with the extension in order to demonstrate its functionality. We show that RDF2Vec~\cite{rdf2vec_journal} embeddings derived directly from the ontologies to be matched are capable of representing the internal structure of an ontology but do not provide any value for matching tasks with differently structured ontologies when evaluated as the only feature. 
We further show that multiple feature generators and a machine learning component help to obtain a high precision alignment in the \emph{Ontology Alignment Evaluation Initiative} (OAEI) \emph{knowledge graph} track~\cite{kg_track_paper_1,kg_track_paper_2}.

\section{Related Work}
\emph{Classification} is a flavor of \emph{supervised learning} and denotes a machine learning approach where the learning system is presented with a set of records carrying a \emph{class} or \emph{label}. Given those records, the system is trained by trying to predict the correct class.~\cite{liu_web_2011_ch3} Transferred to the ontology alignment domain, the set of records can be regarded as a collection of correspondences where some of the correspondences are correct (class \emph{true}) and some correspondences are false (class \emph{false}). Hence, the classification system at hand is binary. 

The application of supervised learning is not new to ontology matching. In fact, even in the very first edition of the OAEI\footnote{Back then the competition was actually referred to as \emph{EON Ontology Alignment Contest}.} in 2004 the \emph{OLA} matching system~\cite{ola_2004} performed a simple optimization of weights using the provided reference alignments. In the past, multiple publications
~\cite{ichise2008machine,DBLP:conf/esws/EckertMS09,shadgara2011ontology,ngo2011generic,pomap_large_paper} addressed supervised learning in ontology matching, occasionally also referred to as \emph{matching learning}. Unsupervised machine learning approaches are less often used, but have been proposed for the task of combining matchers as well \cite{muller2015towards}.

More recently, Nkisi-Orji et al.~\cite{DBLP:conf/pkdd/Nkisi-OrjiWMHH18} present a matching system that uses a multitude of features and a random forest classifier. The system is evaluated on the OAEI \emph{conference} track~\cite{conference_track} and the EuroVoc dataset, but did not participate in the actual evaluation campaign. Similarly, Wang et al.~\cite{wang2018ontology} present a system called \emph{OntoEmma} which exploits a neural classifier together with 32 features. The system is evaluated on the \emph{large biomed} track. However, the system did not participate in an OAEI campaign either. It should be mentioned here that a comparison between systems that have been trained with parts of the reference and systems that have not is not really fair (despite being the typical approach).

Also a recent, OAEI-participating matching system applies supervised learning: The \emph{POMap++} matching system~\cite{pomap_large_paper} uses a local classifier which is not based on the reference alignment but on a locally created gold standard. The system also participated in the last two recent OAEI campaigns~\cite{pomap_oaei_2019,pomap_oaei_2018}. 

The implementations of the approaches are typically not easily reusable or available in a central framework.

\section{The MELT Framework}
\label{sec:the_melt_framework}
\paragraph{Overview} MELT~\cite{melt_reference} is a framework written in Java for ontology and instance matcher development, tuning, evaluation, and packaging. It supports both, HOBBIT and SEALS, two heavily used evaluation platforms in the ontology matching community. The core parts of the framework are implemented in Java, but evaluation and packaging of matchers implemented in other languages is also supported. Since 2020, MELT is the official framework recommendation by the OAEI and the MELT track repository is used to provide all track data required by SEALS. MELT is also capable of rendering Web dashboards for ontology matching results so that interested parties can analyze and compare matching results on the level of correspondences without any coding efforts~\cite{melt_dashboard}. This has been pioneered at the OAEI 2019 for the \emph{knowledge graph} track.\footnote{For a demo of the MELT dashboard, see \url{https://dwslab.github.io/melt/anatomy_conference_dashboard.html}}
MELT is open-source\footnote{\url{https://github.com/dwslab/melt/}}, under a permissive license, and is available on the maven central repository\footnote{\url{https://mvnrepository.com/artifact/de.uni-mannheim.informatik.dws.melt}}.

\paragraph{Different Gold Standard Types}
Matching systems are typically evaluated against a reference alignment. 
A reference alignment may be complete or only partially complete. The latter means that not all entities in the matching task are aligned and that any entity not appearing in the gold standard cannot be judged. Therefore, the following five levels of completeness can be distinguished: (i) complete, (ii) partial with complete target and complete source, (iii) partial with complete target and incomplete source, (iv) partial with complete source and incomplete target, (v) partial with incomplete source and incomplete target. 
If the reference is complete, all correspondences not available in the reference alignment can be regarded as wrong. If only one part of the gold standard is complete (ii, iii, and iv), every correspondence involving an element of the complete side that is not available in the reference can be regarded as wrong. If the gold standard is incomplete (v), the correctness of correspondences not in the gold standard cannot be judged. 
For example, given that the gold standard is partial with complete target and complete source (case ii), and given the correspondence 
$<a,b,=,1.0>$, the correspondence $<a,c,=,1.0>$ could be judged as wrong because it involves $a$ which is from the complete side of the alignment. On the other hand, the correspondence $<d,e,=,1.0>$ cannot be judged because it does not involve any element from the gold standard. This evaluation setting is used for example for the OAEI \emph{knowledge graph} track.
OAEI reference datasets are typically complete with the exception of the \emph{knowledge graph} track. The completeness of references influences how matching systems have to be evaluated. MELT can handle all stated levels of completeness. The completeness can be set for every \texttt{TestCase} separately using the enum \texttt{GoldStandardCompleteness}. The completeness also influences the generation of negative correspondences for a gold standard in supervised learning. MELT supports matching system developers also in this use case.

\section{Supervised Learning Extensions in MELT}

\subsection{Python Wrapper}
\begin{figure}[]
\centering
\includegraphics[scale=0.35]{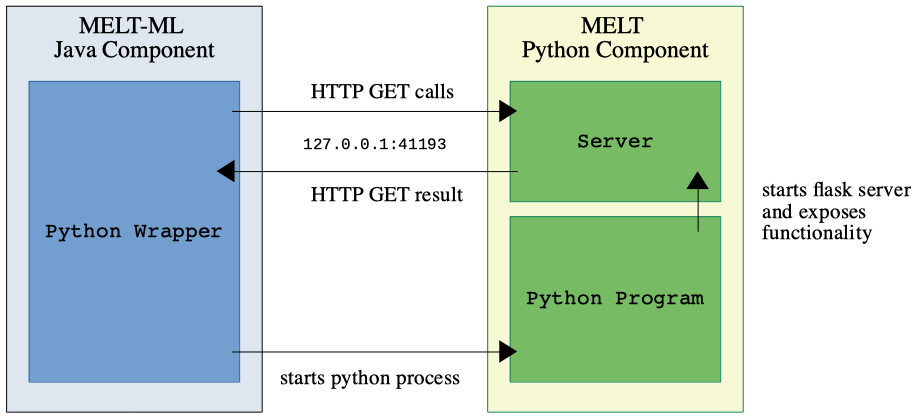}
\caption{Python code execution in MELT.}
\label{fig:python_wrapper}
\end{figure}

As researchers apply advances in machine learning and natural language processing to other domains, they often turn to Python because leading machine learning libraries such as \emph{scikit-learn}\footnote{\url{https://scikit-learn.org/}}, \emph{TensorFlow}\footnote{\url{https://www.tensorflow.org/}}, \emph{PyTorch}\footnote{\url{https://pytorch.org/}}, \emph{Keras}\footnote{\url{https://keras.io/}}, or \emph{gensim}\footnote{\url{https://radimrehurek.com/gensim/}} are not easily available for the Java language. In order to exploit functionalities provided by Python libraries in a consistent manner without a tool break, a wrapper is implemented in MELT which communicates with a Python backend via HTTP as depicted in Figure~\ref{fig:python_wrapper}. The server works out-of-the-box requiring only that Python and the libraries listed in the \texttt{requirements.txt} file are available on the target system. The MELT-ML user can call methods in Java which are mapped to a Python call in the background. As of MELT 2.6, functionality from \emph{gensim} and \emph{scikit-learn} are wrapped.

\subsection{Generation of Training Data}
Every classification approach needs features and class labels.
In the case of matching, each example represents a correspondence and the overall goal is to have an ML model which is capable of deciding if a correspondence is correct or not. Thus, the matching component can only work as a filter e.g. it can only remove correspondences of an already generated alignment.

For training such a classifier, positive and negative examples are required. The positive ones can be generated by a high precision matcher or by an externally provided alignment such as a sample of the reference alignment or manually created correspondences.
As mentioned earlier, no OAEI track provides a dedicated alignment for training. Therefore, MELT provides a new \texttt{sample(int n)} method in the \texttt{Alignment} class for sampling $n$ correct correspondences as well as \texttt{sampleByFraction(double fraction)} for sampling a $fraction$ in range $(0, 1)$ of correct correspondences. 

Negative examples can be easily generated in settings where the gold standard is complete or partially complete (with complete source and/or target, see Section~\ref{sec:the_melt_framework}). The reason is that any correspondence with an entity appearing in the positive examples can be regarded as incorrect. Thus, a recall oriented matcher can generate an alignment and all such correspondences represent the negative class. In cases where the gold standard is partial and the source and/or target is incomplete, each negative correspondence has to be manually created.

\subsection{Generation of Features}
\label{ssec:generation_of_features}
The features for the correspondences are generated by one or more matchers which can be concatenated in a pipeline or any other control flow. MELT provides an explicit framework for storing the feature values in  correspondence extensions (which are by default also serialized in the alignment format). 
The correspondence method \texttt{add\-Additional\-Confidence(String key, double confidence)} is used to add such feature values (more convenience methods exist).

MELT already provides some out-of-the-box feature generators in the form of so called \emph{filters} and \emph{matchers}. A \emph{matcher} detects new correspondences. As of MELT 2.6, 17 matchers are directly available (e.g., different string similarity metrics). A \emph{filter} requires an input alignment and adds the additional confidences to the correspondences, or removes correspondences below a threshold. In MELT, machine learning is also included via a filter (\texttt{Machine\-Learning\-Scikit\-Filter}). As of MELT 2.6, 21 filters are available. A selection is presented in the following:

\paragraph{SimilarNeighboursFilter}
Given an initial alignment of instances, the \texttt{Similar\-Neighbours\-Filter} analyzes for each of the instance correspondences how many already matched neighbours the source and target instances share.
It can be further customized to also include similar literals (defined by string processing methods).
The share of neighbours can be added to the correspondence as absolute value or relative to the total numbers of neighbours for source and target. For the latter, the user can choose from \texttt{min} (size of the intersection divided by minimum number of neighbours of source or target), \texttt{max}, \texttt{jaccard} (size of intersection dived by size of union), and \texttt{dice} (twice the size of intersection divided by the sum of source and target neighbours).

\paragraph{CommonPropertiesFilter}
This filter selects instance matches based on the overlap of properties. The idea is that equal instances also share similar properties. Especially in the case of homonyms, this filter might help. For instance, given two instances with label 'bat', the string may refer to the mammal or to the racket where the first sense has properties like 'taxon', 'age', or 'habitat' and the latter one has properties like 'material', 'quality', or 'producer'. This filter of course requires already matched properties. The added confidence can be further customized similarly to the previous filter. Furthermore, property URIs are by default filtered to exclude properties like \texttt{rdfs:label}.

\paragraph{SimilarHierarchyFilter}
This component analyzes any hierarchy for given instance matches such as type hierarchy or a category taxonomy as given in the \emph{knowledge graph} track. Thus, two properties are needed: 1) instance to hierarchy property which connects the instance to the hierarchy (in case of type hierarchy this is \texttt{rdf:type}) 2) hierarchy property which connects the hierarchy (in case of type hierarchy this is \texttt{rdfs:subClassOf}).
This filter needs matches in the hierarchy which are counted similarly to the previous filters. Additionally, the confidence can be computed by a hierarchy level dependent value (the higher the match in the hierarchy, the lower the confidence). \texttt{SimilarTypeFilter} is a reduced version of it by just looking at the direct parent.

\paragraph{BagOfWordsSetSimilarityFilter}
This filter analyzes the token overlap of the literals given by a specific property. The tokenizer can be freely chosen as well as the overlap similarity.

\paragraph{MachineLearningScikitFilter}
The actual classification part is implemented in class  \texttt{MachineLearningScikitFilter}.
In the standard setting, a five-fold cross validation is executed to search for the model with the best f-measure. The following models and hyper parameters are tested:

\begin{itemize}
	\item \emph{Decision Trees} optimized by minimum leaf size and maximum depth of tree (1-20)
	\item \emph{Gradient Boosted Trees} optimized by maximum depth (1,6,11,16,21) and number of trees (1,21,41,61,81,101)
	\item \emph{Random Forest} optimized by number of trees (1-100 with 10 steps) and minimum leaf size (1-10)
	\item \emph{Na\"{i}ve Bayes} (without specific parameter tuning)
	\item \emph{Support Vector Machines} (SVM) with radial base function kernel; C and gamma are tuned according to~\cite{hsu2003practical}
	\item \emph{Neural Network} with one hidden layer in two different sizes $F/2+2$, $sqrt(F)$, and two hidden layers of $F/2$ and $sqrt(F)$, where $F$ denotes the number of features
\end{itemize}

All of these combinations are evaluated automatically with and without feature normalization (\texttt{MinMaxScaler} which scales each feature to a range between zero and one).
The best model is then trained on the whole training set and applied to the given alignment.

\subsection{Analysis of Matches}
A correspondence which was found by a matching system and which appears in the reference alignment is referred to as \emph{true positive}. A \emph{residual true positive} correspondence is a true positive correspondence that is not trivial as defined by a trivial alignment. The trivial alignment can be given or calculated by a simple baseline matcher. String matches, for instance, are often referred to as trivial. Given a reference alignment, a system alignment, and a trivial alignment, the \emph{residual recall} can be calculated as the share of non trivial correspondences found by the matching system~\cite{aguirre2012results,euzenat_ontology_2013_ch_9}. 

If a matcher was trained using a sample of the reference alignment and is also evaluated on the reference alignment, a true positive match can only be counted as meaningful if it was not available in the training set before. In MELT, the baseline matcher can be set dynamically for an evaluation. Therefore, for supervised matching tasks where a sample from the reference is used, the sample can be set as baseline solution (using the \texttt{ForwardMatcher}) so that only additionally found matches are counted as residual true positives. 
Using the alignment cube file\footnote{The alignment cube file is a CSV file listing all correspondences found and not found (together with filtering properties) that is generated by the \texttt{EvaluatorCSV}.}, residual true positives can be analyzed at the level of individual correspondences.

\section{Exemplary Analysis}

\subsection{RDF2Vec Vector Projections}

\paragraph{Experiment}
In this experiment, the ontologies to be matched are embedded and a projection is used to determine matches. \emph{RDF2Vec} is a knowledge graph embedding approach which generates random walks for each node in the graph to be embedded and afterwards runs the \emph{word2vec}~\cite{word_2_vec_1,word_2_vec_2} algorithm on the generated walks. Thereby, a vector for each node in the graph is obtained. The RDF graph is used in \emph{RDF2Vec} without any pre-processing such as in other approaches like \emph{OWL2Vec}~\cite{owl2vec}.
The embedding approach chosen here has been used on external background knowledge for ontology alignment before~\cite{alod2vec_matcher}.

In this setting, we train embeddings for the ontologies to be matched. In order to do so, we integrate the  \emph{jRDF2Vec}\footnote{\url{https://github.com/dwslab/jRDF2Vec}}~\cite{portisch_rdf2vec_2020} framework into MELT in order to train the embedding spaces. 
Using the functionalities provided in the MELT-ML package, we train a linear projection from the source vector space into the target vector space. In order to generate a training dataset for the projection, the \texttt{sampleByFraction(double fraction)} method is used. 
For each source, the closest target node in the embedding space is determined. If the confidence for a match is above a threshold $t$, the correspondence is added to the system alignment.

Here, we do not apply any additional matching techniques such as string matching. The approach is fully independent of any stated label information. The exemplary matching system is available online as an example.\footnote{\url{https://github.com/dwslab/melt/tree/master/examples/RDF2VecMatcher}}

\paragraph{Results} For the vector training, we generate 100 random walks with a depth of 4 per node and train skip-gram (SG) embeddings with 50 dimensions, minimum count of 1, and a window size of 5. We use a sampling rate of 50\% and a threshold of 0.85. While the implemented matcher fails to generate a meaningful residual recall when the two ontologies to be matched are different, it performs very well when the ontologies are of the same structure as in the \emph{multifarm} track. Here, the approach generates many residual true positives with a residual recall of up to 61\% on \emph{iasted-iasted} as seen in Table~\ref{tab:same-onto-results-rdf2vecprojections}. Thus, it could be shown that \emph{RDF2Vec} embeddings do contain structural information of the knowledge graph that is embedded.

\begin{table}[]
\begin{tabular}{|l|c|c|c|c|c|c|c|}
\hline
\textbf{Multifarm Test Case} & \textbf{P} & \textbf{R} & \textbf{R+} & \textbf{F} & \textbf{\# of TP} & \textbf{\# of FP} & \textbf{\# of FN} \\ \hline
\textbf{iasted-iasted}         & 0.8232 & 0.7459 & 0.6111 & 0.7836 & 135 & 29 & 46 \\ \hline
\textbf{conference-conference} & 0.7065 & 0.5285 & 0.1967 & 0.6047 & 65  & 27 & 58 \\ \hline
\textbf{confOf-confOf}         & 0.9111 & 0.5541 & 0.1081 & 0.6891 & 41  & 4  & 33 \\ \hline
\end{tabular}
\setlength{\belowcaptionskip}{-8pt}
\caption{Performance of RDF2Vec projections on the same ontologies in the multifarm track. \emph{P} stands for \emph{precision}, \emph{r} stands for \emph{recall}, and \emph{R+} for \emph{residual recall}. R+ refers here to the fraction of correspondences found that were previously not available in the training set. \emph{\# of ...} refers to the number of \emph{true positives (TP)}, \emph{false positives (FP)}, and \emph{false negatives (FN)}. Details about the track can be found in~\cite{multifarm}}
\label{tab:same-onto-results-rdf2vecprojections}
\end{table}

\subsection{Knowledge Graph Track Experiments}
\paragraph{Experiment}

In this experiment, the instances of the OAEI \emph{knowledge graph} track are matched.
First, a basic matcher (\texttt{BaseMatcher}) is used to generate a recall oriented alignment by applying simple string matching on 
the property values of \texttt{rdfs:label} and \texttt{skos:altLabel}. The text is compared once using string equality and once in a normalized fashion (non-ASCII characters are removed and the whole string is lowercased).

Given this alignment, the above described feature generators / filters are applied in isolation to re-rank the correspondences and afterwards the \texttt{Naive\-Descending\-Extractor}~\cite{meilicke2007analyzing} is used to create a one-to-one alignment based on the best confidence. 

In contrast to this, another supervised approach is tried out. After executing the \texttt{BaseMatcher}, all feature generators are applied after each other where each filter adds one feature value. The feature values are calculated independently of each other. This results in an alignment where each correspondence has the additional confidences in its extensions. As a last step, the \texttt{Machine\-Learning\-Scikit\-Filter} is executed. The training alignment is generated by sampling all correspondences from the \texttt{Base\-Matcher} where the source \textit{or} target is involved.
The correspondence is a positive training example if the source \textit{and} the target appear in the input alignment (which is in our case the sampled reference alignment) and a negative example in all other cases.

The search for the machine learning model is executed as a five-fold cross validation and the best model is used to classify all correspondences given by the \texttt{BaseMatcher}.
The whole setup is available on GitHub\footnote{\url{https://github.com/dwslab/melt/tree/master/examples/supervisedKGTrackMatcher}}.

\paragraph{Results}
In all filters, the absolute number of overlapping entities are used (they are normalized during a grid search for the best model).
In the \texttt{Similar\-Neigh\-bours\-Filter}, the literals are compared with text equality and the hierarchy filter compares the categories of the Wiki pages. The \texttt{Similar\-Type\-Filter} analyzes the direct classes which are extracted from templates (indicated by the text 'infobox').
The results for this experiment are depicted in Table~\ref{tab:kgResults} which shows that not one feature can be used for all test cases because different Wiki combinations (test cases) require different filters. The \texttt{Base\-Matcher} already achieves a good f-measure which is also in line with previous analyses~\cite{kg_track_paper_3}.
When executing the \texttt{MachineLearningScikitFilter} the precision can be increased for three test cases and the associated drop in recall is relatively small. It can be further seen that there is not one single optimal classifier out of the classifiers tested.
\afterpage{
\clearpage
\begin{landscape}
\begin{table}[t]
\centering
\begin{tabular}{|l|c|c|c|c|c|c|c|c|c|c|c|c|c|c|c|}
    \hline
     & \multicolumn{3}{|c|}{\textbf{mcu-}}   & \multicolumn{3}{|c|}{\textbf{memoryalpha-}} &\multicolumn{3}{|c|}{\textbf{memoryalpha-}} & \multicolumn{3}{|c|}{\textbf{starwars-}} & \multicolumn{3}{|c|}{\textbf{starwars-}}\\
             & \multicolumn{3}{|c|}{\textbf{marvel}} & \multicolumn{3}{|c|}{\textbf{memorybeta}}   & \multicolumn{3}{|c|}{\textbf{stexpanded}}   & \multicolumn{3}{|c|}{\textbf{swg}}       & \multicolumn{3}{|c|}{\textbf{swtor}} \\
    \hline
    \textbf{Approach}  & \textbf{P} & \textbf{R} & \textbf{F} & \textbf{P} & \textbf{R} & \textbf{F} & \textbf{P} & \textbf{R} & \textbf{F} & \textbf{P} & \textbf{R} & \textbf{F} & \textbf{P} & \textbf{R} & \textbf{F}\\
    \hline
    BaseMatcher & 0.8548&0.6796&0.7572&0.8740&0.8978&0.8858&0.8675&0.9264&0.8960&0.9001&0.7318&0.8072&0.9007&0.9146&0.9076\\
    \hline
    CommonPropertiesFilter & 0.8823&0.6614&0.7560&0.9310&0.8785&0.9040&0.9370&0.8968&0.9165&0.9257&0.7162&0.8076&0.9371&0.8999&0.9181 \\
    SimilarHierarchyFilter  & 0.8823&0.6614&0.7560&0.9361&0.8830&0.9088&0.9527&0.9107&0.9312&0.9281&0.7181&0.8097&0.9440&0.9057&0.9245 \\
    BagOfWordsSetSimilarityFilter & 0.8823&0.6614&0.7560&0.9340&0.8810&0.9067&0.9406&0.8991&0.9194&0.9292&0.7190&0.8107&0.9348&0.8976&0.9159 \\
    SimilarNeighboursFilter & 0.8912&0.6687&0.7641&0.9467&0.8916&0.9183&0.9600&0.9171&0.9380&0.9375&0.7254&0.8179&0.9317&0.8947&0.9128 \\
    SimilarTypeFilter   & 0.8823&0.6614&0.7560&0.9247&0.8727&0.8980&0.9303&0.8899&0.9096&0.9222&0.7135&0.8045&0.9326&0.8962&0.9140 \\
    \hline
    ML (sample=0.2)   & 0.8831&0.6620&0.7567&0.9636&0.8592&0.9084&0.9648&0.8887&0.9252&0.9292&0.7190&0.8107&0.9621&0.8778&0.9180 \\
    & \multicolumn{3}{|c|}{SVM} & \multicolumn{3}{|c|}{Random Forest} & \multicolumn{3}{|c|}{SVM} & \multicolumn{3}{|c|}{SVM} & \multicolumn{3}{|c|}{Random Forest}\\ \hline
    ML (sample=0.4)   & 0.8831&0.6620&0.7567&0.9636&0.8599&0.9088&0.9734&0.8690&0.9182&0.9315&0.7199&0.8121&0.9445&0.8903&0.9166\\ 
    & \multicolumn{3}{|c|}{Random Forest} & \multicolumn{3}{|c|}{Random Forest} & \multicolumn{3}{|c|}{Neural Network} & \multicolumn{3}{|c|}{Neural Network} & \multicolumn{3}{|c|}{Random Forest}\\ \hline
    ML (sample=0.6)   & 0.8831&0.6620&0.7567&0.9685&0.8575&0.9096&0.9667&0.8916&0.9276&0.9367&0.7153&0.8112&0.9565&0.8903&0.9222\\ 
    & \multicolumn{3}{|c|}{Random Forest} & \multicolumn{3}{|c|}{Decision Tree} & \multicolumn{3}{|c|}{Neural Network} & \multicolumn{3}{|c|}{SVM} & \multicolumn{3}{|c|}{SVM}\\
    \hline
\end{tabular}
\caption{Precision (P), recall (R), and f-measure (F) for all five test cases of the \emph{knowledge graph} track using different matching approaches. Details about the track can be found in~\cite{kg_track_paper_3}. For the ML approaches, the optimal classifier (given the evaluated ones outlined in Subsection~\ref{ssec:generation_of_features}) is stated below the scores.}
\label{tab:kgResults}
\end{table}
\end{landscape}
}

\section{Conclusion and Outlook}
With MELT-ML, we have presented a machine learning extension for the MELT framework which facilitates feature generation and feature combination. The latter are included as \emph{filters} to refine existing matches. MELT also allows for the evaluation of ML-based matching systems.

In the future, we plan to extend the provided functionality by the Python wrapper to further facilitate machine learning in matching applications. We further plan to extend the number of feature generators. With our contribution we hope to encourage OAEI participants to apply and evaluate supervised matching techniques. In addition, we intend to further study different strategies and ratios for the generation of negative examples.

We further would like to emphasize that a special machine learning track with dedicated training and testing alignments might benefit the community, would increase the transparency in terms of matching system performance, and might further increase the number of participants since researchers use OAEI datasets for supervised learning but there is no official channel to participate if parts of the reference alignment are required. 

\bibliography{references}
\end{document}